\documentclass{article}
\usepackage{ijcai19}

\usepackage{times}
\usepackage{soul}
\usepackage{url}
\usepackage[draft]{hyperref} % Yifan, otherwise raise an error
\usepackage[utf8]{inputenc}
\usepackage[small]{caption}
\usepackage{graphicx}
\usepackage{amsmath}
\usepackage{booktabs}
\usepackage{algorithm}
\usepackage{algorithmic}
\usepackage{amsmath}
\usepackage{amssymb}
\usepackage{multirow}
\usepackage{array}
\usepackage{graphicx}
\usepackage{comment}
\usepackage{booktabs}
\usepackage{subfigure}
\usepackage{caption}
\usepackage{xcolor,colortbl} 
\usepackage{multirow}
\usepackage{makecell}

\newcommand{\citet}[1]{\citeauthor{#1}~\shortcite{#1}}

\title{Difficulty Controllable Generation of Reading Comprehension Questions}

\author{
Yifan Gao$^1$\thanks{This work was partially done when Yifan Gao was an intern at Tencent AI Lab working with Lidong Bing, who was a full-time researcher there.}
\and
Lidong Bing$^2$
\and
Wang Chen$^{1}$\and
Michael R. Lyu$^1$\And
Irwin King$^1$
\affiliations
$^1$Department of Computer Science and Engineering,\\ The Chinese University of Hong Kong, Shatin, N.T., Hong Kong\\
$^2$R\&D Center Singapore, Machine Intelligence Technology, Alibaba DAMO Academy
\emails
$^1$\{yfgao,wchen,lyu,king\}@cse.cuhk.edu.hk,
$^2$l.bing@alibaba-inc.com
}
\begin{document}

\maketitle

\begin{abstract}
We investigate the difficulty levels of questions in reading comprehension datasets such as SQuAD, and propose a new question generation setting, named \textbf{D}ifficulty-controllable \textbf{Q}uestion \textbf{G}eneration (\textbf{DQG}). Taking as input a sentence in the reading comprehension paragraph and some of its text fragments (i.e., answers) that we want to ask questions about, a DQG method needs to generate questions each of which has a given text fragment as its answer, and meanwhile the generation is under the control of specified difficulty labels---the output questions should satisfy the specified difficulty as much as possible.
To solve this task, we propose an end-to-end framework to generate questions of designated difficulty levels by exploring a few important intuitions.
For evaluation, we prepared the first dataset of reading comprehension questions with difficulty labels. The results show that the question generated by our framework not only have better quality under the metrics like BLEU, but also comply with the specified difficulty labels. 
\end{abstract}

\section{Introduction}
\begin{figure}
\centering
\includegraphics[width=1.0\columnwidth]{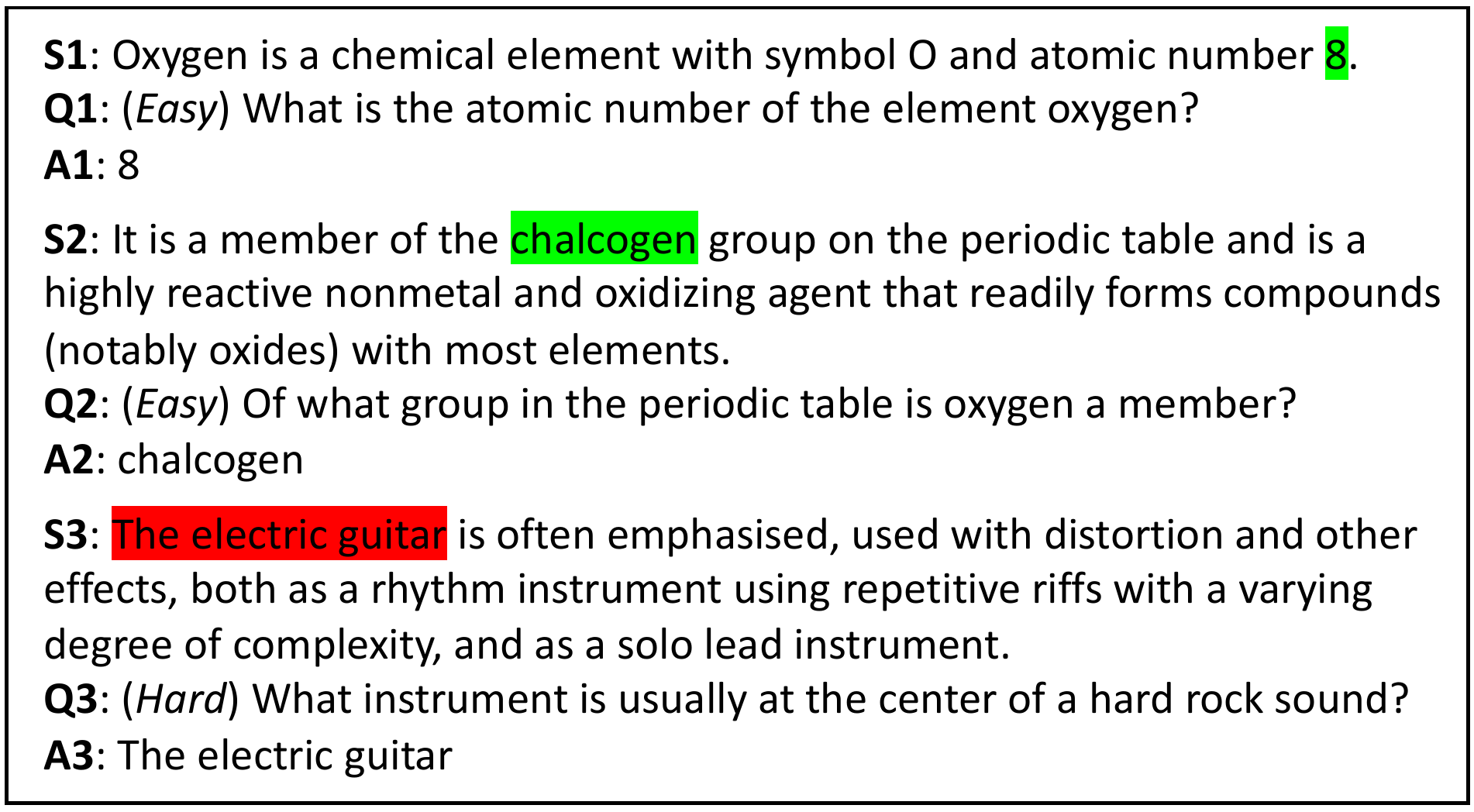}
\caption{\label{figure:SQuAD_demo}Example questions from SQuAD. The answers of Q1 and Q2 are facts described in the sentences, thus they are easy to answer. But it is not straightforward to answer Q3}
\end{figure}

% generally introduce qg
Question Generation (QG) aims to generate natural and human-like questions from a range of data sources, such as image \cite{Mostafazadeh2016GeneratingNQ}, knowledge base \cite{Serban2016GeneratingFQ,Su2016OnGC}, and free text \cite{Du2017LearningTA}. Besides for constructing SQuAD-like dataset \cite{Rajpurkar2016SQuAD10}, QG is also helpful for the intelligent tutor system: The instructor can actively ask the learner questions according to reading comprehension materials \cite{Heilman2010GoodQS} or particular knowledge  \cite{Danon2017ASA}. 
In this paper, we focus on QG for reading comprehension text. For example, Figure \ref{figure:SQuAD_demo} gives three questions from SQuAD, our goal is to generate such questions.

% introduce papers in qg for mc
QG for reading comprehension is a challenging task because the generation should not only follow the syntactic structure of questions, but also ask questions to the point, i.e., having a specified aspect as its answer. Some template-based approaches \cite{Vanderwende2007AnsweringAQ,Heilman2010GoodQS} were proposed initially, where well-designed rules and heavy human labor are required for declarative-to-interrogative sentence transformation. 
With the rise of data-driven learning approach and sequence to sequence (seq2seq) framework, some researchers formulated QG as a seq2seq problem  \cite{Du2017LearningTA}: The question is regarded as the decoding target from the encoded information of its corresponding input sentence. 
However, different from existing seq2seq learning tasks such as machine translation and summarization which could be loosely regarded as learning a one-to-one mapping, for question generation, different aspects of the given descriptive sentence can be asked, and hence the generated questions could be significantly different.
Several recent works tried to tackle this problem by incorporating the answer information to indicate what to ask about, which helps the models  generate more accurate questions \cite{Song2018LeveragingCI,Zhou2017NeuralQG}. 
In our work, we also focus on the answer-aware QG problem, which assumes the answer is given. Similar problems have been addressed in, e.g., \cite{Zhao2018ParagraphlevelNQ,Sun2018AnswerfocusedAP}.

% introduce our difficulty-aware setting
In this paper, we investigate a new setting of QG, namely \textbf{D}ifficulty controllable \textbf{Q}uestion \textbf{G}eneration (\textbf{DQG}). In this setting, given a sentence in the reading comprehension paragraph, the text fragments (i.e., answers) that we want to ask questions about, and the specified difficulty levels, a framework needs to generate questions that are asked about the specified answers and satisfy the difficulty levels as much as possible. For example, given the sentence S3 and the answer ``the electric guitar'' in Figure \ref{figure:SQuAD_demo}, the system should be capable of asking both a hard question like Q3 and an easy one such as ``What is often emphasised as a rhythm instrument?''. 
DQG has rich application scenarios. For instance, when instructors prepare learning materials for students, they may want to balance the numbers of hard questions and easy questions. 
Besides, the generated questions can be used to test how a QA system works for questions with diverse difficulty levels.

Generating questions with designated difficulty levels is a more challenging task. First, no existing large-scale QA dataset has difficulty labels for questions to train a reliable neural network model. Second, for a single sentence and answer pair, we want to generate questions with diverse difficulty levels. However, the current datasets like SQuAD only have one given question for each sentence and answer pair. Finally, there is no metric to evaluate the difficulty of questions.
To overcome the first issue, we prepare a dataset of reading comprehension questions with difficulty labels. Specifically, we design a method to automatically label SQuAD questions with multiple difficulty levels, and obtain 76K questions with difficulty labels.

To overcome the second issue, we propose a framework that can learn to generate questions complying with the specified difficulty levels by exploring the following intuitions. To answer a SQuAD question, one needs to locate a text fragment in the input sentence as its answer. Thus, if a question has more hints that can help locate the answer fragment, it would be easier to answer. For the examples in Figure \ref{figure:SQuAD_demo}, the hint ``atomic number'' in Q1 is very helpful, because, in the corresponding sentence, it is just next to the answer ``8'', while for Q3, the hint ``instrument'' is far from the answer ``The electric guitar''. 
The second intuition is inspired by the recent research on style-guided text generation, which incorporates a latent style representation (e.g., sentiment label or review rating score) as an input of the generator \cite{DBLP:conf/nips/ShenLBJ17,quase}. Similarly, performing difficulty control can be regarded as a problem of sentence generation towards a  specified attribute or style. On top of the typical seq2seq architecture, our framework has two tailor-made designs to explore the above intuitions: (1) Position embeddings are learned to capture the proximity hint of the answer in the input sentence; 
(2) Global difficulty variables are learned to control the overall ``difficulty'' of the questions. For the last issue, we propose to employ the existing reading comprehension (RC) systems to evaluate the difficulty of generated questions. Intuitively, questions which cannot be answered by RC systems are more difficult than these correctly answered ones.

In the quantitative evaluation, we compare our DQG model with state-of-the-art models and ablation baselines. The results show that our model not only generates questions of better quality under the metrics like BLEU and ROUGE, but also has the capability of generating questions complying with the specified difficulty labels. The manual evaluation finds that the language quality of our generated questions is better, and our model can indeed control the question difficulty.

\section{Task Definition}
In the DQG task, our goal is to generate SQuAD-like questions of diverse difficulty levels for a given sentence. Note that the answers of SQuAD questions are text spans in the input sentence, and they are significantly different from RACE questions \cite{lai2017large} such as ``What do you learn from the story?''. Considering their different emphases, SQuAD questions are more suitable for our task, while the difficulty of RACE questions mostly comes from the understanding of the story but not from the way how the question is asked.
Thereby, we assume that the answers for asking questions are given, and they appear as text fragments in the input sentences by following the paradigm of SQuAD. 

We propose an end-to-end framework to handle DQG. Formally, let $a$ denote the answer for asking question, let $s$ denote the sentence containing $a$ from a reading comprehension paragraph.
Given $a$, $s$, and a specified difficulty level $d$ as input, the DQG task is to generate a question $q$ which has $a$ as its answer, and meanwhile should have $d$ as its difficulty level. 

\section{The Protocol of Difficulty Labeling}

SQuAD \cite{Rajpurkar2016SQuAD10} is a reading comprehension dataset containing 100,000+ questions on Wikipedia articles. The answer of each question is a text fragment from the corresponding input passage. 
We employ SQuAD questions to prepare our experimental dataset. 

The difficulty level is a subjective notion and can be addressed in many ways, e.g., syntax complexity, coreference resolution and elaboration \cite{Sugawara2017EvaluationMF}. 
To avoid the ambiguity of the ``question difficulty'' in this preliminary study, we design the following automatic labeling protocol and study the correlation between automatically labelled difficulty with human difficulty. 
We first define two difficulty levels, \textit{Hard} and \textit{Easy}, in this preliminary dataset for the sake of simplicity and practicality. We employ two RC systems, namely R-Net \cite{NETWORKS2017RnetMR}~\footnote{\scriptsize{\url{https://github.com/HKUST-KnowComp/R-Net}}} and BiDAF \cite{Seo2016BidirectionalAF}~\footnote{\scriptsize{\url{https://github.com/allenai/bi-att-flow}}}, to automatically assess the difficulty of the questions. The \textit{labeling protocol} is: A question would be labelled with \textit{Easy} if both R-Net and BiDAF answer it correctly under the exact match metric, and labelled with \textit{Hard} if both systems fail to answer it. The remaining questions are eliminated for suppressing the ambiguity. 

Note that we cannot directly employ the original data split of SQuAD to train a model of R-Net or BiDAF, and use the model to assess all questions. Such assessment is not appropriate, because models will overfit training questions and label them all as easy ones. To avoid this problem, we re-split the original SQuAD questions into 9 splits and adopt a 9-fold strategy. To label every single split (the current split), 7 splits are used as the training data, and the last split is used as the validation data. Then the trained model is used to assess the difficulty of questions in the current split. This way guarantees that the model is never shown with the questions for automatic labeling.
Finally, we obtain 44,723 easy questions and 31,332 hard questions. 

To verify the reasonability of our labeling protocol, we evaluate its consistency with human being's judgment. 
We sample 100 \textit{Easy} questions and 100 \textit{Hard} questions, and hire 3 annotators to rate the difficulty level of all these questions on a 1-3 scale (3 for the most difficult). The result shows that average difficulty rating for the \textit{Easy} questions is 1.90 while it is 2.52 for the \textit{Hard} ones.

\begin{figure}[t!]
\centering
\includegraphics[width=1.0\columnwidth]{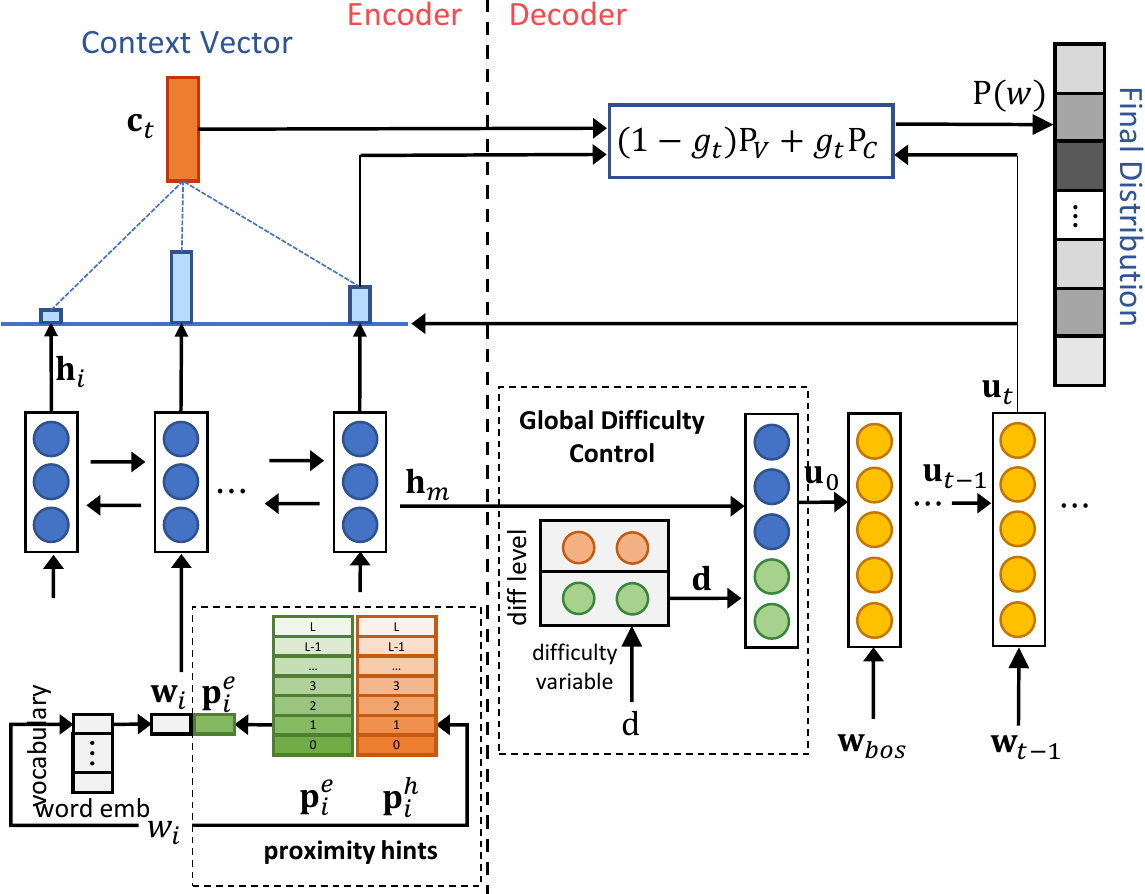}
\caption{Overview of our DQG framework \textit{(better viewed in color)}}
\label{figure:pipeline}
\end{figure}

\section{Framework Description}
Given an input sentence $s=({w_1}, {w_2}, ..., {w_m})$, a text fragment $a$ in $s$, and a difficulty level $d$, our task is to generated a question $q$, which is asked with $s$ as its background information, takes $a$ as its answer, and has $d$ as its difficulty. 
The architecture of our difficulty-controllable question generator is depicted in Figure \ref{figure:pipeline}. The encoder takes two types of inputs, namely, the word embeddings and the relative position embeddings (capturing the proximity hints) of sentence words (including the answer words). Bidirectional LSTMs are employed to encode the input into contextualized representations. Besides two standard elements, namely attention and copy, the decoder contains a special initialization to control the difficulty of the generated question. Specifically, we map the difficulty label $d$ into a global difficulty variable with a lookup table, and combine the variable with the last hidden state of the encoder to initialize the decoder.

\subsection{Exploring Proximity Hints}
Recall that our first intuition tells that the proximity hints are helpful for answering the SQuAD-like questions. Before introducing our design for implementing the intuition, we quantitatively verify it by showing some statistics. Specifically, we examine the average distance of those nonstop question words that also appear in the input sentence to the answer fragment. For example, for Q1 in Figure \ref{figure:SQuAD_demo} and its corresponding input sentence ``Oxygen is a chemical element with symbol O and atomic number 8'', we calculate the word-level average distance of words ``atomic'', ``number'', ``element'', and ``oxygen'' to the answer ``8''. The statistics are given in Table~\ref{tab:distance_stat}.
In contrast, the average distance of all nonstop sentence words to the answer is also given in the bottom line. If we only count those nonstop question words, we find that their distance to the answer fragment is much smaller than the sentence words, namely 8.43 vs. 11.20. We call this \textit{Question Word Proximity Hint }(\textbf{QWPH}). More importantly, the distance for hard questions is significantly larger than that for easy questions, namely 9.71 vs. 7.67, which well verifies our intuition that if a question has more obvious proximity hints (i.e., containing more words that are near the answer in the corresponding sentence), it would be easier to solve. We model QWPH for easy questions and hard questions separately and call this \textit{Difficulty Level Proximity Hint }(\textbf{DLPH}).

To implement the QWPH intuition, our model learns a lookup table which maps the distance of each sentence word to the answer fragment, i.e., $0$ (for answer words), 1, 2, etc., into a position embedding: $(\mathbf{p}_0, \mathbf{p}_{1}, \mathbf{p}_{2}, ..., \mathbf{p}_{L})$, where $\mathbf{p}_i\in{\mathbb{R}^{d_p}}$ and $d_p$ is the dimension. $L$ is the maximum distance we consider.
Different from QWPH that is difficulty agnostic, the DLPH intuition additionally explores the information of question difficulty levels. Therefore, we define two lookup tables: $(\mathbf{p}_0^e, \mathbf{p}_{1}^e, \mathbf{p}_{2}^e, ..., \mathbf{p}_{L}^e)$ for the Easy label, and $(\mathbf{p}_0^h, \mathbf{p}_{1}^h, \mathbf{p}_{2}^h, ..., \mathbf{p}_{L}^h)$ for the Hard label. 
Note that the above position embeddings not only carry the information of sentence word position, but also let our model know which aspect (i.e., answer) to ask with the embeddings of position 0.

\begin{table}[t!]
    \small
    \centering
    {\small
    \resizebox{1.0\columnwidth}{!}{
    \begin{tabular}{l|ccc} 
    \Xhline{2\arrayrulewidth}
    \hline
    & {~Easy}   & {~Hard}     & {~All}     \\ 
    \hline
    {Avg. distance of question words}   & {~~7.67}   & {~~9.71}    & {~~8.43}     \\ 
    \hline
    {Avg. distance of all sentence words}   & {11.23}   & {11.16}    & {11.20}     \\ 
    \hline
    \Xhline{2\arrayrulewidth}
    \end{tabular}}
    }
    \caption{Distance statistics for non-stop words}
    \label{tab:distance_stat}
\end{table}

\subsection{Characteristic-rich Encoder} 
The characteristic-rich encoder incorporates several features into a contextualized representation.
For each sentence word $w$, an embedding lookup table is firstly used to map tokens in the sentence into dense vectors: $(\mathbf{w}_1,$ $\mathbf{w}_2, ..., \mathbf{w}_m)$, where $\mathbf{w}_i\in{\mathbb{R}^{d_w}}$ of $d_w$ dimensions. 
Then we concatenate its word embedding and position embedding (proximity hints) to derive a characteristic-rich embedding: $\mathbf{x}=[\mathbf{w};\mathbf{p}$]. 
We use bidirectional LSTMs to encode the sequence $(\mathbf{x}_1, \mathbf{x}_2, ..., \mathbf{x}_m)$ to get a contextualized representation for each token:
\begin{equation}\label{eqn.enc}
\overrightarrow{\mathbf{h}}_i = \overrightarrow{\text{{LSTM}}}(\overrightarrow{\mathbf{h}}_{i-1}, \mathbf{x}_i),~ \overleftarrow{\mathbf{h}}_i = \overleftarrow{\text{{LSTM}}}(\overleftarrow{\mathbf{h}}_{i+1}, \mathbf{x}_i), \nonumber
\end{equation}
where $\overrightarrow{\mathbf{h}}_i$ and $\overleftarrow{\mathbf{h}}_i$ are the hidden states at the  $i$-th time step of the forward and the backward LSTMs. 
We concatenate them together as $\mathbf{h}_i=[\overrightarrow{\mathbf{h}}_i;\overleftarrow{\mathbf{h}}_i]$. 

\subsection{Difficulty-controllable Decoder}
We use another LSTM as the decoder to generate the question. We employ the difficulty label $d$ to initialize the hidden state of the decoder.
During the decoding, we incorporate the attention and copy mechanisms to enhance the performance.

\paragraph{Global Difficulty Control. }
We regard the generation of difficulty-controllable questions as a problem of sentence generation towards a specified style, i.e., easy or hard.
To do so, we introduce a global difficulty variable to control the generation. We follow the recent works for the task of style transfer that apply the control variable globally, i.e., using the style variable to initialize the decoder \cite{quase}.  
Specifically, for the specified difficulty level $d$, we first map it to its corresponding difficulty variable $\mathbf{d}\in{\mathbb{R}^{d_d}}$, where $d_d$ is the dimension of a difficulty variable. Then we use the concatenation of $\mathbf{d}$ with the final hidden state $\mathbf h_m$ of the encoder to initialize the decoder hidden state $\mathbf{u}_0 = [\mathbf h_m;\mathbf{d}]$.
Note that in the training stage, we feed the model the ground truth difficulty labels, while in the testing stage, our model can take any specified difficulty labels, i.e., difficulty-controllable, for question generation.
We have also tried some variations by adding this variable to other places such as every encoder or decoder input in the model but it does not work.

\paragraph{Decoder with Attention \& Copy. }
The decoder predicts the word probability distribution at each decoding timestep to generate the question.
At the \textit{t}-th timestep, it reads the word embedding $\mathbf{w}_{t}$ and the hidden state $\mathbf{u}_{t-1}$ of the previous timestep to generate the current hidden state $\mathbf u_t = \text{LSTM}(\mathbf u_{t-1}, \mathbf w_{t})$.
Then the decoder employs the attention mechanism \cite{Luong2015EffectiveAT,Jiani18,Wang2019TitleguidedEF} and copy mechanism \cite{See2017GetTT} to generate the question by copying words in the sentence or generating words from a predefined vocabulary.

% \subsection{Training and Inference}
% In the training, our model minimizes the following negative log-likelihood of all training instances:
% \begin{equation}\label{eqn.cr_qg_train}
% \mathcal{L} =  - \sum_{q \in \mathcal{Q}} \text{log P(}q|{a}, s, d\text{)},
% \end{equation}
% where $\mathcal{Q}$ includes all training data points, and $\text{log P(}q|{a}, s, d\text{)}$ is the conditional log-likelihood of $q$.
% For testing, we can generate questions of diverse difficulty levels $d_i\in\mathcal{D}$ (predefined difficulty levels) by maximizing:
% \begin{equation}\label{eqn.cr_qg}
% \overline{q} = \argmax_{q} \text{log P(}q|a, s, d_i\text{)}.
% \end{equation}

\section{Experiments}
\subsection{Experimental Settings}
\paragraph{Dataset.}
Our prepared dataset is split according to articles of the SQuAD data, and Table \ref{tab:stat} provides the detailed statistics.
Across the training, validation and test sets, the splitting ratio is around 7:1:1, and the easy sample ratio is around 58\% for all three.

\begin{table}[!t]
\small
    \centering
    {\small
    \resizebox{0.7\columnwidth}{!}{
    \begin{tabular}{l|ccc} 
    \Xhline{2\arrayrulewidth}
    \hline
    & {Train}   & {Dev}     & {Test}     \\ 
    \hline
    {\# easy questions}   & {34,813}   & {4,973}    & {4,937}     \\ 
    \hline
    {\# hard questions}   & {24,317}   & {3,573}    & {3,442}     \\ 
    \hline
    {Easy ratio} & {58.88\%} & {58.19\%} & {58.92\%}  \\
    \hline
    \Xhline{2\arrayrulewidth}
    \end{tabular}}
    }
    \caption{The statistics of our dataset}
    \label{tab:stat}
\end{table}

\paragraph{Baselines and Ablation Tests.}
We only employ neural network based methods as our baselines, since they perform better than non-neural methods as shown in recent works \cite{Du2017LearningTA,Zhou2017NeuralQG}.
The first baseline models the question generation as a seq2seq problem incorporating the attention mechanism, and we refer to it as \textbf{L2A} \cite{Du2017LearningTA}. The second baseline
\textbf{Ans} adds answer indicator embedding to the seq2seq model, similar to \cite{Zhou2017NeuralQG,Kumar2018AutomatingRC}.
Two ablations that only employ the question word proximity hint or the difficulty level proximity hint are referred to as \textbf{QWPH} and \textbf{DLPH}.
Moreover, we examine the effectiveness of the global difficulty control (\textbf{GDC}) combined with QWPH and DLPH, refer to them as \textbf{QWPH-GDC} and \textbf{DLPH-GDC}. All these methods are enhanced by the \textit{copy} mechanism. 

\paragraph{Model Details and Parameter Settings.}
The embedding dimensions for the position embedding and the global difficulty variable, i.e. $d_p$ and $d_d$, are set to 50 and 10 respectively. We use the maximum relative distance $L=20$ in the position embedding.
We adopt teacher-forcing in the encoder-decoder training and use the ground truth difficulty labels. 
In the testing procedure, we select the model with the lowest perplexity and beam search with size 3 is employed for question generation.
All important hyper-parameters, such as $d_p$ and $d_d$, are selected on the validation dataset.

\begin{table}[!t]
    \small
    \centering
    \resizebox{0.95\columnwidth}{!}{
    \begin{tabular}{l | c c c c | c c c c}
    \Xhline{3\arrayrulewidth}
    & \multicolumn{4}{c|}{\textbf{Easy} Questions Set} & \multicolumn{4}{c}{\textbf{Hard} Questions Set} \\
    & \multicolumn{2}{c}{R-Net} & \multicolumn{2}{c|}{BiDAF} & \multicolumn{2}{c}{R-Net} & \multicolumn{2}{c}{BiDAF} \\  
  & \texttt{EM} & \texttt{F1} & \texttt{EM} & \texttt{F1} &  \texttt{EM} &  \texttt{F1} & \texttt{EM} & \texttt{F1} \\
    \hline
    \hline
   Ans & 82.16 & 87.22 & 75.43  & 83.17 & 34.15 & 60.07 & 29.36 & 55.89 \\
   QWPH & 82.66 & 87.37 & 76.10 & 83.90 & 33.35 & 59.50 & 28.40 & 55.21 \\
   QWPH-GDC & 84.35 & 88.86 & 77.23 & 84.78 & 31.60 & 57.88 & 26.68 & 54.31 \\
   DLPH & 85.49 & 89.50 & 78.35 & 85.34 & 28.05 & 54.21 & 24.89 & 51.25 \\ 
   DLPH-GDC & \textbf{85.82} & \textbf{89.69} & \textbf{79.09} & \textbf{85.72} & \textbf{26.71} & \textbf{53.40} & \textbf{24.47} & \textbf{51.20} \\
    \Xhline{3\arrayrulewidth}
    \end{tabular}}
    \caption{Difficulty of the generated questions, measured with R-Net and BiDAF. For easy questions, higher score indicates better difficulty-control, while for hard questions, lower indicates better}
    \label{tab:generation-result-auto}
\end{table}

\begin{table}[!t]
    \centering
    \resizebox{0.95\columnwidth}{!}{
    \begin{tabular}{l | c c c c | c c c c}
    \Xhline{3\arrayrulewidth}
    & \multicolumn{4}{c|}{\textbf{Easy} Questions Set} & \multicolumn{4}{c}{\textbf{Hard} Questions Set} \\  
    & \multicolumn{2}{c}{R-Net} & \multicolumn{2}{c|}{BiDAF} & \multicolumn{2}{c}{R-Net} & \multicolumn{2}{c}{BiDAF} \\  
   & \texttt{EM} & \texttt{F1}  & \texttt{EM} & \texttt{F1} &  \texttt{EM} &  \texttt{F1} & \texttt{EM} & \texttt{F1} \\
    \hline
    \hline
   QWPH-GDC & 7.41 & 5.72 & 7.13 & 5.88 & 6.45 & 5.47 & 6.13 & 5.10 \\
   DLPH & 12.41 & 9.51 & 11.28 & 8.49 & 12.01 & 10.45 & 10.51 & 9.37 \\ 
   DLPH-GDC & \textbf{12.91} & \textbf{9.95} & \textbf{12.40} & \textbf{9.23} & \textbf{12.68} & \textbf{10.76} & \textbf{11.22} & \textbf{9.97} \\
    \Xhline{3\arrayrulewidth}
    \end{tabular}}
    \caption{The results of controlling difficulty, measured with R-Net and BiDAF. The scores are performance gap between questions generated with original difficulty label and questions generated with reverse difficulty label}
    \label{tab:generation-gap-auto}
\end{table}

\subsection{Difficulty Control Results}

We run R-Net and BiDAF to assess the difficulty of our generated hard and easy questions. Here the R-Net and BiDAF systems are trained using the same train/validation splits as shown in Table \ref{tab:stat}, and we report their performance under the standard reading comprehension measures for SQuAD questions, i.e., Exact Match (\textbf{EM}) and macro-averaged F1 score (\textbf{F1}), on the easy and hard question sets respectively.
For all experiments, we firstly show the performance of difficulty-controllable question generation by feeding ground truth difficulty labels, then we feed the reverse difficulty labels to demonstrate our model can \textit{\textbf{control}} the difficulty of generated questions. 

Recall that the generated questions can be split into an easy set and a hard set according to the difficulty labels. Here we evaluate the generated questions from the perspective that a reading comprehension system (e.g., R-Net and BiDAF) should perform better on the generated questions in the easy set, and perform worse on the hard question set. If a pipeline does not use the answer information, its generated questions are likely not about the answers, thus both BiDAF and R-Net cannot work well no matter for easy or hard questions. Therefore, we do not use L2A here. 

\begin{table}[!t]
    \centering
    \resizebox{0.8\columnwidth}{!}
    {\small
    \begin{tabular}{l| c c c| c c c}
    \Xhline{3\arrayrulewidth}
    \multirow{2}{*}{\textbf{}} & \multicolumn{3}{c|}{\textbf{Easy} Question Set} & \multicolumn{3}{c}{\textbf{Hard} Question Set} \\ \cline{2-7}
    & \texttt{F} & \texttt{D} & \texttt{R} & \texttt{F} & \texttt{D} & \texttt{R} \\ \hline
    Ans           & 2.91     & 2.02        & 0.74      & 2.87     & 2.12        & 0.58      \\
    DLPH-GDC           & 2.94     & 1.84        & 0.76      & 2.87     & 2.26        & 0.64     \\ 
    \Xhline{3\arrayrulewidth}
    \end{tabular}}
    \caption{Human evaluation results for generated questions. \texttt{Fluency(F)} and \texttt{Difficulty(D)} take values from \{1, 2, 3\} (3 means the top fluency or difficulty), while \texttt{Relevance(R)} takes a binary value, i.e., 1 or 0}
    \label{tab:manual}
\end{table}

As shown in Table \ref{tab:generation-result-auto}, for the easy set, the questions generated by the methods using the difficulty label ``Easy'' are easier to answer. Specifically, compared with Ans and QWPH which cannot control the difficulty, QWPH-GDC, DLPH, and DLPH-GDC generate easier questions, showing that they have the capability of generating difficulty-controllable questions. One instant doubt is that a model can simply produce trivial questions by having them contain the answer words. In fact, our models do not have this behaviour, because it will increase the training loss. To further verify this, we calculate the occurrence rate of answer words in the generated questions. The result shows that only 0.09\% answer words appear in the questions generated by our models.

For the hard set, we can draw the same conclusion by keeping in mind that a lower score indicates the corresponding method performs better in generating difficulty-controllable questions. (Note that questions irrelevant to the answer can also yield lower scores, and we have more discussion about this issue in Section \ref{sec:human} for the human evaluation.)
This observation shows that incorporating the difficulty information locally by the two position embeddings or globally by the difficulty-controlled initialization indeed guides the generator to generate easier or harder questions. 
Comparing DLPH and QWPH-GDC, we find that the local difficulty control by the position embedding is more effective. DLPH-GDC performs the best by combining the local and global difficulty control signals. 

Moreover, we find that QWPH achieves slightly better performance than Ans baseline. 
A large performance gap between QWPH-GDC and QWPH again validates the effectiveness of the global difficulty control. 
Meanwhile, the improvement from QWPH to DLPH shows that the local difficulty level proximity hint can stress the question difficulty at each time step to perform better.

On the other hand, another way to validate our model is testing whether our model can \textit{\textbf{control}} the difficulty by feeding the reversed difficulty labels.
For example, for a question in the easy set, if we feed the ``Hard'' label together with the input sentence and answer of this question into our model, we expect the generated question should be harder than feeding the ``Easy'' label. 
Concretely, if a method has the better capability in controlling the difficulty, on two sets of questions generated with this method by taking the true label and the reversed label, the performance gap of a reading comprehension system should be larger. The results of this experiment are given in Table \ref{tab:generation-gap-auto}.
We only compare models which have difficulty control capability. 
The model combining local and global difficulty signals, i.e., DLPH-GDC, achieves the largest gap, which again shows that: (1) DLPH-GDC has the strongest capability of generating difficulty-controllable questions; (2) The local difficulty control (i.e. DLPH) is more effective than the global (i.e. QWPH-GDC).

\subsection{Manual Evaluation} \label{sec:human}
% basic information
%Since neither N-gram metrics like BLEU nor reading comprehension pipelines are perfect to evaluate the generated questions completely, here we conduct a human evaluation. 
We hire 3 annotators to rate the model generated questions. We randomly sampled 100 question with ``Easy`` labels and 100 with ``Hard`` labels from the test set, and let each annotator annotate these 200 cases. 
During the annotation, each data point contains a sentence, an answer, and the questions generated by different models, without showing the difficulty labels.
%In this section, we conduct a human evaluation, which is a necessity for question generation and particularly true for our DQG task. 
We consider three metrics: \texttt{Fluency(F)}, \texttt{Difficulty(D)} and \texttt{Relevance(R)}. The annotators are first asked to read the generated questions to evaluate their grammatical correctness and fluency. Then, all annotators are required to rate the difficulty of each generated question by considering the corresponding sentence and answer. Finally, for relevance, we ask the annotators to judge if the question is asking about the answer. 
\texttt{Fluency} and \texttt{Difficulty} take values from \{1, 2, 3\} (3 means the top fluency or difficulty), while \texttt{Relevance} takes a binary value (1 or 0). 

% result analysis
Table \ref{tab:manual} shows the results of the manual evaluation. We compare our best model DLPH-GDC with the Ans baseline. We separate the \texttt{Easy} questions and \texttt{Hard} questions for statistics.
For both question sets, both models achieve high scores on \texttt{Fluency}, owing to the strong language modeling capability of neural models. 
For \texttt{Difficulty}, we can find that DLPH-GDC can generate easier or harder questions than Ans by feeding the true difficulty labels. 
Another observation is that, for the Ans baseline, questions generated in the \texttt{Easy} set are easier than those in the \texttt{Hard} set, which validates our difficulty labelling protocol from another perspective. Note that for human beings, all SQuAD-like questions are not really difficult, therefore, the difference of \texttt{Difficulty} values between the easy set and the hard set is not large.  

Furthermore, we can observe our model can generate more relevant questions compared with the Ans baseline. The reason could be that our position embedding can not only tell where the answer words are, but also indicate the distance of the context words to the answer. Thus, it provides more information to the model for asking to the point questions. Ans only differentiates the answer token and non-answer token, and treats all non-answer tokens equally. 

Recall that we had the concern regarding Table \ref{tab:generation-result-auto} that the generated hard questions by our difficulty-controlling models say DLPH-GDC may simply be irrelevant to the answer, which makes DLPH-GDC achieves lower EM/F1 scores than the Ans baseline. By comparing the \texttt{Relevance} scores in Table \ref{tab:manual} and EM/F1 scores in Table \ref{tab:generation-result-auto} for Hard Question Set, we find that the questions generated by DLPH-GDC are more relevant (as shown in Table \ref{tab:manual}) and more difficult (as shown in both Tables \ref{tab:generation-result-auto} and \ref{tab:manual}) than those generated by the Ans baseline. This observation resolves our doubt on the irrelevance issue and supports the conclusion that our DLPH-GDC does generate more difficult and relevant questions which can fail the two RC pipelines.

\subsection{\mbox{Automatic Evaluation of Question Quality}}

Here we evaluate the similarity of generated questions with the ground truth. Since our dataset is not parallel (i.e., for a sentence and answer pair, our dataset only has one question with the ``easy'' or ``hard'' label), here we only evaluate the question quality by feeding the ground truth difficulty labels.
We employ BLEU (B), METEOR (MET) and ROUGE-L (R-L) scores by following \cite{Du2017LearningTA}. BLEU evaluates the average N-gram precision on a set of reference sentences, with a penalty for overly long sentences. ROUGE-L is commonly employed to evaluate the recall of the longest common subsequences, with a penalty for short sentences.

\begin{table}[t]
    \small
    \centering
    \resizebox{0.80\columnwidth}{!}{
    \begin{tabular}{@{}l@{~} | @{~}c@{~} @{~}c@{~} @{~}c@{~} @{~}c@{~} @{~}c@{~} @{~}c@{~} }
    \Xhline{3\arrayrulewidth}
    \hline
    & B1 & B2 & B3 & B4 & MET & R-L \\
      \hline
    \hline
    L2A & 36.01 & 21.61 & 14.97 & 10.88 & 15.99 & 38.06 \\
    Ans & 43.51 & 29.06 & 21.35 & 16.22 & 20.53 & 45.66  \\
    QWPH  & 43.75 & 29.28 & 21.61 & 16.46 & 20.70 & 46.02 \\
    \cline{1-7}
    QWPH-GDC & 43.99 & 29.60 & 21.86 & 16.63 & 20.87 & 46.26 \\
    DLPH & 44.11 & 29.64 & 21.89 & 16.68 & 20.94 & 46.22  \\
    DLPH-GDC & 43.85 & 29.48 & 21.77 & 16.56 & 20.79 & 46.16  \\
    
    \Xhline{3\arrayrulewidth}
    
    \end{tabular}}
    \caption{Automatic evaluation for question quality}
    \label{tab:generation-result-sim}
\end{table}

Table \ref{tab:generation-result-sim} shows the quality of generated questions.
Comparing the first three methods, we can find that the answer and position information helps a lot for asking to the point questions, i.e., more similar to the ground truth. Moreover, QWPH performs better than Ans, indicating that further distinguishing the different distance of the non-answer words to the answer provides richer information for the model to generate better questions. 
The results in the lower half show that, given the ground truth difficulty labels, these three methods with the capability of difficulty control are better than the first three methods.
These three models achieve comparable performance, and DLPH-GDC sacrifices a little in N-gram based performance here while achieving the best difficulty control capability (refer to Tables \ref{tab:generation-result-auto} \& \ref{tab:generation-gap-auto}).

\subsection{Case Study}
\begin{figure}
\centering
\includegraphics[width=0.8\columnwidth]{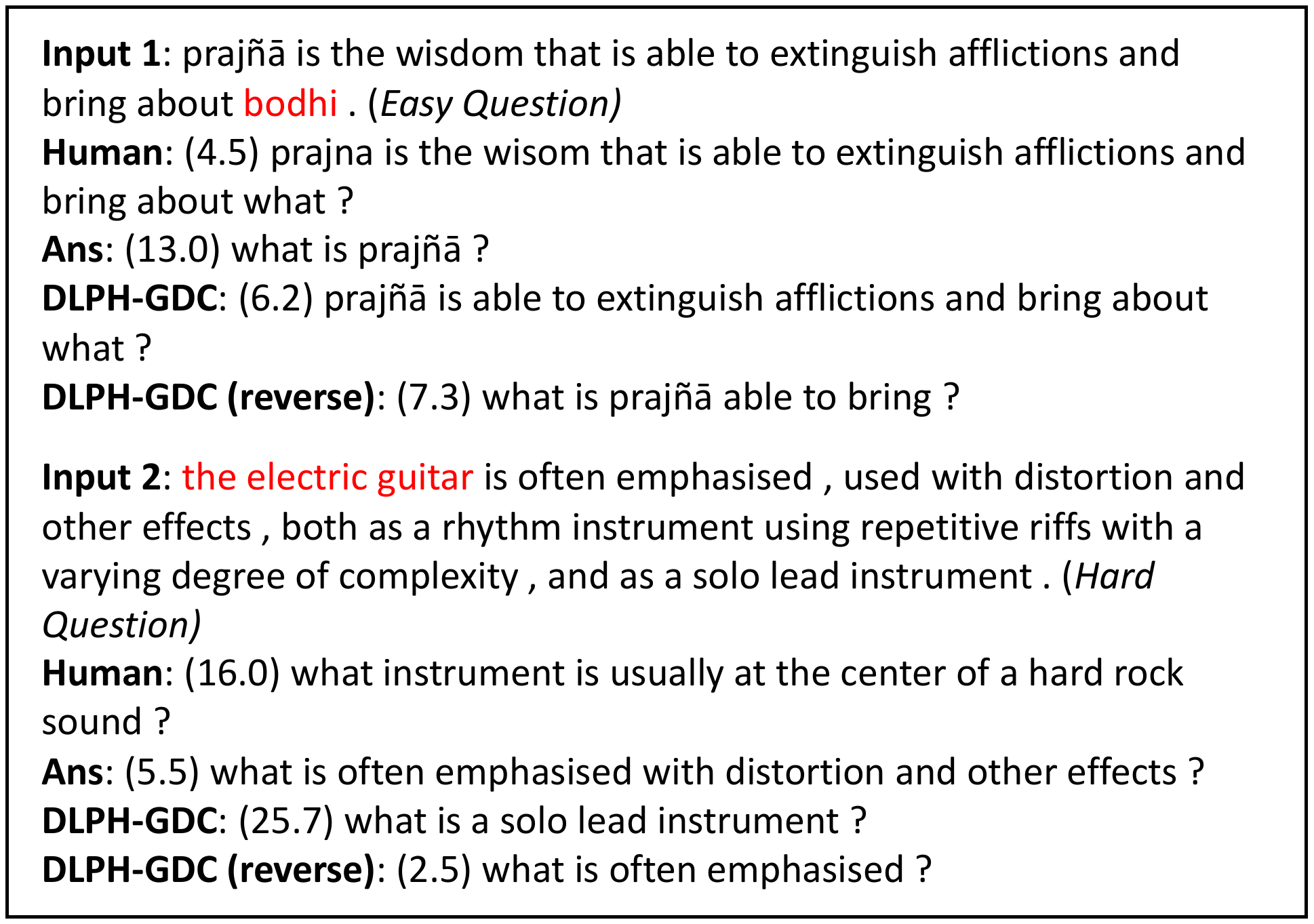}
\caption{Example questions (with answers marked in red). The human question for Input 2 uses some information (``hard rock'') in preceding sentences which are not shown here} 
\label{figure:casestudy}
\end{figure}

Figure \ref{figure:casestudy} provides some examples of generated questions (with answers marked in red). The number after the model is the average distance of the overlapped nonstop words between the question and the input sentence to the answer fragment. The average distance corresponds to the our intuition proximity hints well. Compared with questions generated by Ans baseline, our model can give more hints (shorter distance) when asking easier questions and give less hints (longer distance) when asking harder questions.

For the first example, we observe that the ground truth question generated by Human is quite easy, just replacing the answer ``bodhi'' with ``what''. Among the three systems, Ans asks a question that is not about the answer. While both DLPH-GDC and DLPH-GDC (reverse) are able to generate to the point questions. Specifically, by taking the ``Easy'' label, DLPH-GDC tends to use more words from the input sentence, while DLPH-GDC (reverse) uses less and its generated question is relatively difficult. 
For the second example, we find our system is also applicable to the question with ``Hard'' label.

\section{Related Work}\label{sec.lit_review}

In this section, we primarily review question generation (QG) works on free text. 
\citet{Vanderwende2007AnsweringAQ} proposed this task, 
later on, several rule-based approaches were proposed. They manually design some question templates and transform the declarative sentences to interrogative questions \cite{Mazidi2014LinguisticCI,Labutov2015DeepQW,Lindberg2013GeneratingNL,Heilman2010GoodQS}. 
These Rule-based approaches need extensive human labor to design question templates, and usually can only ask annotators to evaluate the generated questions.

% learning based
\citet{Du2017LearningTA} proposed the first automatic QG framework. They view QG as a seq2seq learning problem to learn the mapping between sentences and questions in reading comprehension. 
Moreover, the procedure of QG from a sentence is not a one-to-one mapping, because given a sentence, different questions can be asked from different aspects. As \citet{Du2017LearningTA} mentioned, in their dataset, each sentence corresponds to 1.4 questions on average.
Seq2seq learning may not perform well for learning such a one-to-many mapping.
% sentence level
Some recent works attempt to solve this issue by assuming the aspect has been already known when asking a question \cite{Zhou2017NeuralQG,Yuan2017MachineCB} or can be detected with a third-party pipeline \cite{Du2018HarvestingPQ}.
This assumption makes sense, because for humans to ask questions, we usually first read the sentence to decide which aspect to ask.
In this paper, we explore another important dimension in QG, i.e., generating questions with controllable difficulty, that has never been studied before.

\section{Conclusions}

In this paper, we present a novel setting, namely difficulty-controllable question generation for reading comprehension, which to the best of our knowledge has never been studied before. 
We propose an end-to-end approach to learn the question generation with designated difficulty levels.
We also prepared the first dataset for this task, and extensive experiments show that our framework can solve this task reasonably well. 
One interesting future direction is to explore generating multiple questions for different aspects in one sentence \cite{Gao2019GeneratingDF}.

\section*{Acknowledgments}

This work is supported by the Research Grants Council of the Hong Kong
Special Administrative Region, China (No. CUHK 14208815 and No. CUHK
14210717 of the General Research Fund). 
We thank Department of Computer Science and Engineering, The Chinese University of Hong Kong for the conference grant support.
We would like to thank Jianan Wang for her efforts in the preliminary investigation.

\bibliographystyle{named}
\bibliography{ijcai19}

\end{document}